\documentclass[10pt,twocolumn,letterpaper]{article}

\usepackage{cvpr}
\usepackage{times}
\usepackage{epsfig}
\usepackage{graphicx}
\usepackage{amsmath}
\usepackage{amssymb}
\usepackage{subcaption}
\usepackage[export]{adjustbox}
\usepackage{multicol}
\usepackage{color}
\usepackage{multirow}
\newcommand{\bs}[1]{\boldsymbol{\mathbf{#1}}}
\newcommand{\secc}[1]{\autoref{sec:#1}}

\usepackage[breaklinks=true,bookmarks=false]{hyperref}

\cvprfinalcopy % *** Uncomment this line for the final submission

\def\cvprPaperID{1385} % *** Enter the CVPR Paper ID here

\ifcvprfinal\fi
\begin{document}

%%%%%%%%% TITLE
\title{Unsupervised Pixel--Level Domain Adaptation\\ with Generative Adversarial Networks}

\author{Konstantinos Bousmalis\\ %Nathan Silberman, David Dohan, Dumitru Erhan, Dilip Krishnan\\
Google Brain\\
London, UK\\
{\tt\small konstantinos@google.com}
\and
Nathan Silberman\\
Google Research\\
New York, NY\\
{\tt\small nsilberman@google.com}
\and
David Dohan\thanks{Google Brain Residency Program: \url{g.co/brainresidency}.}\\
Google Brain\\
Mountain View, CA\\
{\tt\small ddohan@google.com}\and
Dumitru Erhan\\
Google Brain\\
San Francisco, CA\\
{\tt\small dumitru@google.com}\and
Dilip Krishnan\\
Google Research\\
Cambridge, MA\\
{\tt\small dilipkay@google.com}
}

\begin{multicols}{2}
\maketitle
\end{multicols}
%\thispagestyle{empty}

%%%%%%%%% ABSTRACT
\begin{abstract}

Collecting well-annotated image datasets to train modern machine learning algorithms is prohibitively expensive for many tasks. An appealing alternative is to render synthetic data where ground-truth annotations are generated automatically. Unfortunately, models trained purely on rendered images often fail to generalize to real images. To address this shortcoming, prior work introduced unsupervised domain adaptation algorithms that attempt to map representations between the two domains or learn to extract features that are domain--invariant. In this work, we present a new approach that learns, in an unsupervised manner, a transformation in the pixel space from one domain to the other.
Our generative adversarial network (GAN)--based model adapts source-domain images to appear as if drawn from the target domain. Our approach not only produces plausible samples, but also outperforms the state-of-the-art on a number of unsupervised domain adaptation scenarios by large margins. Finally, we demonstrate that the adaptation process generalizes to object classes unseen during training.

\end{abstract}

%%%%%%%%% BODY TEXT

\section{Introduction}

\begin{figure}[t]
\centering
    \begin{subfigure}[b]{.48\textwidth}
        \centering
        \includegraphics[width=\linewidth]{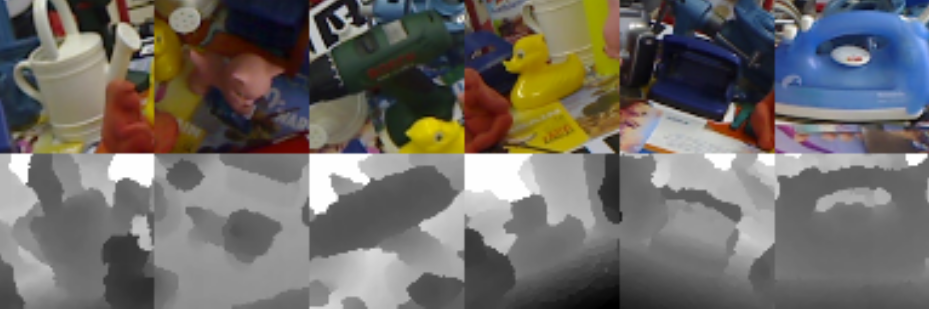}
        \caption{Image examples from the Linemod dataset.}
    \end{subfigure}
    \begin{subfigure}[b]{.48\textwidth}
        \centering
        \includegraphics[width=\linewidth]{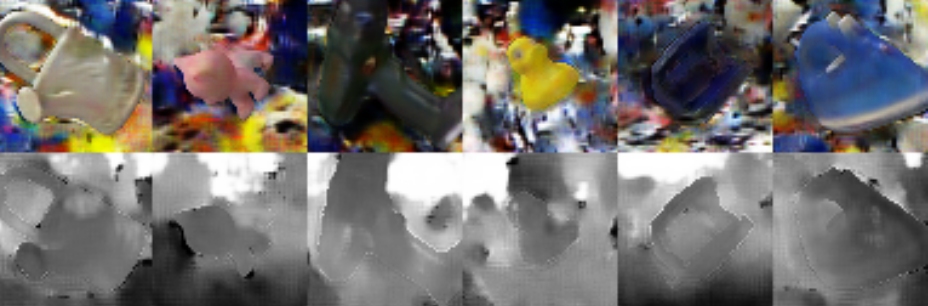}
        \caption{Examples generated by our model, trained on Linemod.}
    \end{subfigure} 
    \quad
\caption{RGBD samples generated with our model vs real RGBD samples from the Linemod dataset~\cite{hinterstoisser2012accv,wohlhart2015learning}. In each subfigure the top row is the RGB part of the image, and the bottom row is the corresponding depth channel. Each column corresponds to a specific object in the dataset. See Sect.~\ref{sec:experiments} for more details.}
\label{fig:teaser}
\end{figure}

Large and well--annotated datasets such as ImageNet
\cite{imagenet_cvpr09}, COCO \cite{lin2014microsoft} and Pascal VOC \cite{everingham2015pascal} 
are considered crucial to advancing computer vision research. However, creating such datasets is 
prohibitively expensive.
One alternative is the use of synthetic data for model training.
It has been a long-standing goal in computer vision to use game engines or
renderers to produce virtually unlimited
quantities of labeled data.
Indeed, certain areas of research, such as deep reinforcement learning for robotics tasks,
effectively require that models be trained in synthetic domains as training in real--world environments can be excessively expensive~\cite{rusu2016sim, tzeng2015towards}. Consequently, there has been a
renewed interest in training models in the synthetic domain and applying them in real--world 
settings \cite{christiano2016transfer, zhu2016target, rusu2016sim, tzeng2015towards, johnson2016driving, mahendran2016researchdoom, qiu2016unrealcv, richter2016playing}.
Unfortunately, models naively trained on synthetic data do not
typically generalize to real images.

A solution to this problem is using unsupervised domain adaptation. In this setting,
we would like to transfer knowledge learned from a source domain, for which we have
labeled data, to a target domain for which we have no labels.
Previous work either attempts to find a mapping from representations of the source domain
to those of the target~\cite{sun2015return}, or seeks to find domain-invariant representations
that are shared between the two domains
~\cite{ganin2016domain,tzeng2015simultaneous,long2015learning,bousmalis2016domain}. While such approaches
have shown good progress, they are still not on par with purely supervised
approaches trained only on the target domain.

In this work, we train a model to
change images from the source domain to appear
as if they were sampled from the target domain while maintaining their original
content. We propose a novel Generative Adversarial Network (GAN)--based architecture
that is able to learn such a transformation in an unsupervised manner, i.e. without using
corresponding pairs from the two domains. 
Our unsupervised pixel-level domain adaptation method (PixelDA) offers a number of advantages over existing approaches:
\vspace{-0.5cm}
\paragraph{Decoupling from the Task-Specific Architecture:} In most 
domain adaptation approaches, the process of domain adaptation and the task-specific
architecture used for inference  
are tightly integrated. One cannot switch a task--specific component of
the model without having to re-train the entire domain
adaptation process. In contrast, because our PixelDA model maps one image to 
another at the pixel level, we can
alter the task-specific architecture without having to re-train the domain adaptation component.

\vspace{-0.5cm}
\paragraph{Generalization Across Label Spaces:} Because previous models couple domain
adaptation with a specific task, the label spaces in the source and target domain
are constrained to match. In contrast, our PixelDA model is able to handle cases where the target label space at test time differs from the label space at training time.

\vspace{-0.5 cm}
\paragraph{Training Stability:} Domain adaptation approaches that rely on some form of adversarial training~\cite{bousmalis2016domain,ganin2016domain} are  sensitive to random initialization. To address this, we incorporate a task--specific loss trained on both source and generated images and a pixel similarity regularization that allows us to avoid mode collapse~\cite{salimans2016improved} and stabilize training. By using these tools, we are able to
reduce variance of performance for the same hyperparameters across different random initializations of our model (see \secc{experiments}). 
%Note that incorporating the task--specific loss at training time doesn't affect our ability to use a different task--specific classifier or a different label space during test time.

\vspace{-0.5cm}
\paragraph{Data Augmentation:} Conventional domain adaptation approaches are limited
to learning from a finite set of source and target data. However, by
conditioning on both source images and a stochastic noise vector, our model 
can be used to create virtually unlimited stochastic samples that appear similar
to images from the target domain. 

\vspace{-0.5cm}
\paragraph{Interpretability:} The output of PixelDA, a domain--adapted image, is much more easily 
interpreted than a domain adapted feature vector.

To demonstrate the efficacy of our strategy, we focus on the tasks of object
classification and pose estimation, where the
object of interest is in the foreground of a given image, for both source and
target domains. Our method outperforms the state-of-the-art unsupervised domain adaptation
techniques on a range of datasets for object classification and pose estimation, 
while generating images that look very similar to the target domain (see \autoref{fig:teaser}).

\section{Related Work}
\label{sec:related_work}

Learning to perform unsupervised domain adaptation is an open theoretical and practical problem. While much prior work exists, our literature 
review focuses primarily on Convolutional Neural Network (CNN) methods due
to their empirical superiority on the problem~\cite{ganin2016domain, long2015learning, sun2015return, tzeng2015ddc}.

\vspace{-0.5cm}
\paragraph{Unsupervised Domain Adaptation:}

Ganin \etal ~\cite{ganin2014unsupervised,ganin2016domain} and Ajakan
\etal ~\cite{Ajakan2014} introduced the Domain--Adversarial Neural
Network (DANN): an architecture trained to extract domain-invariant features.
Their model's first few layers are shared by two classifiers: the first
predicts task-specific class labels when provided with source data
while the second is trained to predict the domain of its inputs.
DANNs minimize the domain classification loss with respect to parameters
specific to the domain classifier, while maximizing it with respect to
the parameters that are common to both classifiers. This minimax optimization becomes possible in a single step via the use of a gradient reversal layer. While
DANN's approach to domain adaptation is to make the features extracted from both domains similar, our approach is to adapt the source images to look as if they
were drawn from the target domain.
Tzeng \etal \cite{tzeng2015ddc} and Long \etal
\cite{long2015learning} proposed versions of DANNs where
the maximization of the domain classification loss is replaced by the minimization
of the Maximum Mean Discrepancy (MMD) metric \cite{gretton2012mmd},
computed between features extracted from sets of samples from each
domain. 
Ghifary \etal \cite{ghifary2016deep} propose an alternative model in which the task loss for the source domain is combined with a reconstruction loss for the target domain, which results in learning domain-invariant features.
Bousmalis \etal \cite{bousmalis2016domain} introduce a model
that explicitly separates the
components that are private to each domain from those that are common 
to both domains. They make use of a
reconstruction loss for each domain, a similarity loss (eg. DANN, MMD) which
encourages domain invariance, and a difference loss which encourages
the common and private representation components to be complementary. 

Other related techniques involve learning a mapping from one domain to the other at a feature level. In such a setup, the feature extraction pipeline is fixed during the domain adaptation optimization. 
This has been applied in various non-CNN based approaches
\cite{gong2012geodesic,caseiro2015beyond,gopalan2011domain} as well as the more recent
CNN-based Correlation Alignment (CORAL) \cite{sun2015return} algorithm.% which
% ``recolors'' whitened source features with the covariance of features from the
% target domain. 

\vspace{-0.5cm}
\paragraph{Generative Adversarial Networks:}

Our model uses GANs
\cite{goodfellow2014generative} conditioned on source images and noise
vectors.
Other recent works have also attempted to use GANs conditioned on 
images. Ledig \etal \cite{ledig2016photo} used an image-conditioned GAN for super-resolution. 
Yoo \etal \cite{yoo2016pixel} introduce the
task of generating images of clothes from images of models wearing them, by 
training on \textit{corresponding pairs} of the clothes worn by models and on a 
hanger. In contrast to our work, neither method conditions on both images and noise vectors, and ours is also applied to an entirely
different problem space.

The work perhaps most similar to ours is that of Liu and Tuzel
\cite{liu2016coupled} who introduce an architecture of a pair of coupled GANs, one for the source and one for the target domain,
whose generators share their high-layer weights and whose discriminators
share their low-layer weights. In this manner, they are able to generate
corresponding pairs of images which can be used for unsupervised domain adaptation. %The success of this approach, however, is contingent
on the ability to generate high quality samples from noise alone. 

\vspace{-0.5cm}
\paragraph{Style Transfer:} The popular work of Gatys \etal \cite{gatys2015neural, gatys2016image}
introduced a method of \textit{style transfer}, in which the style of 
one image is transferred to another while holding the content fixed. The process requires backpropagating back
to the pixels. Johnson \etal \cite{johnson2016perceptual} introduce
a model for feed forward style transfer. They train a network conditioned
on an image to produce an output image whose activations on a pre-trained
model are similar to both the input image (high-level content activations)
and a single target image (low-level style activations). However, both of these approaches are optimized to replicate the
style of a \emph{single} image as opposed to our work which seeks to
replicate the style of an \emph{entire domain} of images.

\section{Model}
\label{sec:model}

\begin{figure*}
\includegraphics[width=\textwidth]{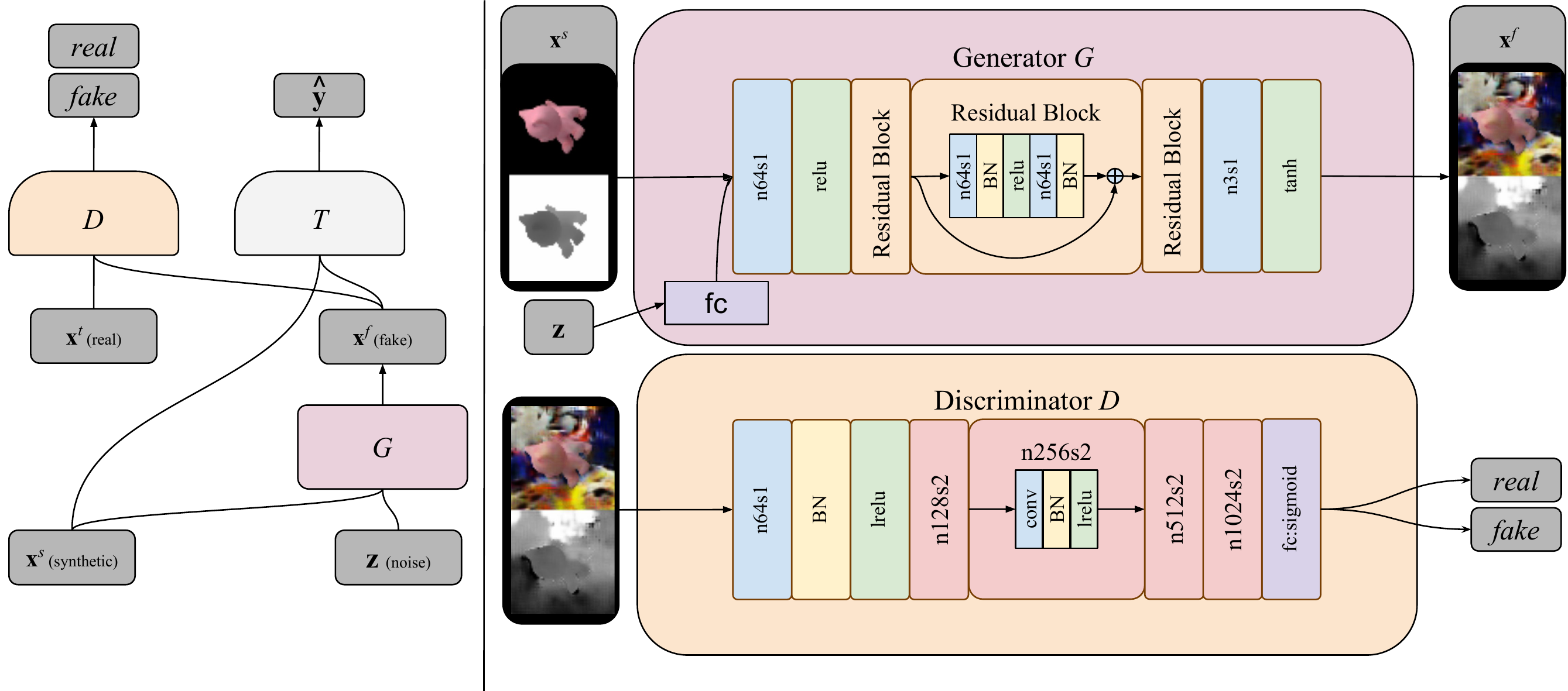}
\caption{An overview of the model architecture.  On the left, we depict the overall model architecture following the style in \cite{odenaACGAN}. On the right, we expand the details of the generator and the discriminator components. The generator $G$ generates an image conditioned on a synthetic image ${\bs x}^s$ and a noise vector $\bs z$. The discriminator $D$ discriminates between real and fake images. The task--specific classifier $T$ assigns task--specific labels $\bs y$ to an image. A convolution with stride 1 and 64 channels is indicated as n64s1 in the image. lrelu stands for leaky ReLU nonlinearity. BN stands for a batch normalization layer and FC for a fully connected layer.  Note that we are not displaying the specifics of $T$ as those are different for each task and decoupled from the domain adaptation process. }
\centering
\label{fig:arch}
%\vspace{-9 pt}
\end{figure*}

We begin by explaining our model for unsupervised pixel-level domain adaptation (PixelDA) in the context of image classification, though our method is not specific to this particular task.
Given a labeled dataset in a source domain and an unlabeled dataset in a target domain, our goal is to train a classifier on data from the source domain that generalizes to the target domain. 
Previous work  performs this task using a single network that performs both domain adaptation and image classification, making the domain adaptation process specific to the classifier architecture. Our model decouples the process of domain adaptation from the process of task-specific classification, as its primary function is to adapt images from the source domain to make them appear as if they were sampled from the target domain. Once adapted, any off-the-shelf classifier can be trained to perform the task at hand as if no domain adaptation were required.
Note that we assume that the differences between the
domains are primarily low-level (due to noise, resolution, illumination, color) rather than high-level (types of objects, geometric variations, etc).

More formally, let ${\bs X}^s = \{{\bs x}_i^s, {\bs y}_i^s\}_{i=0}^{N^s}$
represent a labeled dataset of $N^s$ samples from the source domain and let
${\bs X}^t = \{{\bs x}_i^t\}_{i=0}^{N^t}$ represent an unlabeled
dataset of $N^t$ samples from the target domain.
Our pixel adaptation model consists of a generator function
$G({\bs x}^s, {\bs z}; {\bs \theta}_G) \rightarrow {\bs x^f}$,
parameterized by ${\bs \theta}_G$, that maps a source domain image
${\bs x}^s \in {\bs X}^s$ and a noise vector ${\bs z} \sim p_z$
to an adapted, or fake, image ${\bs x^f}$. Given
the generator function $G$, it is possible to create a new dataset
${\bs X}^f = \left\{ G({\bs x}^s, {\bs z}), {\bs y}^s\right\}$ of any size.
Finally, given an adapted dataset ${\bs X}^f$, the task-specific
classifier can be trained as if the training and test data were 
from the same distribution.

\subsection{Learning}
\label{sec:learning}

To train our model, we employ a generative adversarial objective 
to encourage $G$ to produce images that are similar to the target domain images. During training, our generator
$G({\bs x}^s, {\bs z}; {\bs \theta}_G) \rightarrow {\bs x^f}$
maps a source image ${\bs x}^s$ and a noise vector ${\bs z}$
to an adapted image ${\bs x^f}$. Furthermore, the model is
augmented by a discriminator function $D({\bs x};{\bs \theta}_D)$ that outputs the likelihood $d$ that a given image $\bs x$ has been sampled from the target domain. The discriminator tries to distinguish between `fake' images ${\bs X}^f$ produced by the generator, and `real' images from the target domain ${\bs X}^t$.
Note that in
contrast to the standard GAN formulation \cite{goodfellow2014generative}
in which the generator is conditioned only on a noise vector, 
our model's
generator is conditioned on \emph{both}  a noise vector and an image
from the source domain. In addition to the discriminator, the model is also augmented with a classifier $T({\bs x};{\bs \theta}_T) \rightarrow {\hat{\bs y}}$ which assigns task-specific labels ${\hat{\bs y}}$ to
images ${\bs x} \in \{{\bs X}^f, {\bs X}^t\}$.

Our goal is to optimize the following minimax objective:
\begin{equation}
\min_{{\bs \theta}_G, {\bs \theta}_T} \max_{{\bs \theta}_D}  \;  \alpha \, {\cal L}_{d}(D,G)  + \beta{\cal L}_{t}(G, T)
\label{eq:objective1}
\end{equation}
where $\alpha$ and $\beta$ are weights that control the interaction of the losses.
${\cal L}_{d}$ represents the domain loss:
\begin{align}
{\cal L}_{d}(D, G) = & \; \mathbb{E}_{{\bs x}^t} [\log D({\bs x}^t; {\bs \theta}_D)] + \nonumber \\  & \; \mathbb{E}_{{\bs x}^s, {\bs z}  }[\log (1 - D(G({\bs x^s}, {\bs z}; {\bs \theta}_G); {\bs \theta}_D))]
\end{align}
${\cal L}_{t}$ is a task-specific loss, and in the case of classification we use a typical softmax cross--entropy loss:
\begin{align}
\label{eq:task_loss}
{\cal L}_{t}(G, T) \; = \; \mathbb{E}_{{\bs x}^s,{\bs y}^s, {\bs z}}\big[&-\bs {y^s}^\top \log T\left(G({\bs x^s},
{\bs z}; {\bs \theta}_G); {\bs \theta}_T\right) \nonumber\\
&-\bs {y^s}^\top \log {T({\bs x}^s); {\bs \theta}_T}\big]
\end{align}
where $\bs{y}^s$ is the one-hot encoding of the class label for source input ${\bs x}^s$. 
Notice that we train $T$ with both adapted and non-adapted source images.
When training $T$ only on adapted images, it's possible to achieve
similar performance, but doing so may require many runs with different 
initializations due to the instability of the model. Indeed, without training
on source as well, the model is free to shift class assignments (e.g. class 1 becomes 2, class 2 becomes 3 etc) while still being successful at optimizing the training objective. We have found that training classifier $T$ on
\emph{both} source and adapted images avoids this scenario and greatly stabilizes training (See Table~\ref{tab:stability}). %Finally, it's important to reiterate that once trained, we are free to adapt other images from the source domain which
might use a different label space (See Table~\ref{tab:label_space_exp}).

In our implementation, $G$ is a convolutional neural network with residual 
connections that maintains the resolution of the original image as illustrated in
figure \ref{fig:arch}. Our discriminator $D$ is also a convolutional neural 
network. The minimax optimization of \autoref{eq:objective1} is achieved by alternating between two steps. During the first step, we update the discriminator and task-specific parameters ${\bs \theta}_D, {\bs \theta}_T$, while keeping the generator parameters ${\bs \theta}_G$ fixed. During the second step we fix ${\bs \theta}_D, {\bs \theta}_T$ and update ${\bs \theta}_G$.

\subsection{Content--similarity loss}

In certain cases, we have prior knowledge regarding the low-level
image adaptation process. For example, we may expect the hues of the source
and adapted images to be the same. In our case, for some of our experiments, we render single objects
on black backgrounds and consequently we expect images adapted from these
renderings to have similar foregrounds and different backgrounds from the equivalent source images. Renderers typically provide
access to z-buffer masks that allow us to differentiate between
foreground and background pixels.
This prior knowledge can be formalized via the use of an
additional loss that penalizes large differences between source and
generated images for foreground pixels only. 
Such a similarity loss grounds the generation process to the original image and helps stabilize the minimax optimization, as shown in Sect.~\ref{sec:model_analysis} and Table~\ref{tab:stability}. Our optimization objective then becomes:
\begin{equation}
\min_{{\bs \theta}_G, {\bs \theta}_T} \max_{{\bs \theta}_D}  \;  \alpha{\cal L}_{d}(D,G)  + \beta{\cal L}_{t}(T,G) + \gamma{\cal L}_{c}(G)
\label{eq:objective2}
\end{equation}
where $\alpha$, $\beta$, and $\gamma$ are weights that control the interaction of the losses, and ${\cal L}_c$ is the content--similarity loss.  

A number of losses could anchor the generated image to the original image in some meaningful way (e.g. L1, or L2 loss, similarity in terms of the activations of a pretrained VGG network). In our experiments for learning object instance classification from rendered images, we use a \textit{masked pairwise mean squared error}, which is a variation of the pairwise mean squared error (PMSE) \cite{eigen2014depth}. This loss
penalizes differences between pairs of pixels rather than absolute differences between inputs and outputs. Our masked version calculates the PMSE between the generated foreground and the source foreground. Formally, given a binary mask ${\bs m} \in {\mathbb R}^k$, our masked-PMSE loss is:
\begin{align}
\label{eq:reconstruction_loss}
\mathcal{L}_{c}(G) &= \mathbb{E}_{{\bs x}^s, {\bs z}} \Big[
\frac{1}{k} \left\|\left({\bs x}^s - G({\bs x^s}, {\bs z}; {\bs \theta}_G)  \right ) \circ {\bs m}\right\|_2^2  \nonumber \\
& - \frac{1}{k^2}\left(({\bs x}^s - G({\bs x^s}, {\bs z}; {\bs \theta}_G) )^\top {\bs m}\right)^2 \Big]
\end{align} % {\bs x}, \hat{{\bs x}}
where  $k$ is the number of pixels in input $\bs x$,  $\|\cdot\|_2^2$ is the squared $L_2$-norm, and $\circ$ is the Hadamard product. This loss allows the model to learn to reproduce the overall shape of the objects
being modeled without wasting modeling power on the absolute color or intensity
of the inputs, while allowing our adversarial training to change the object in a consistent way. Note that the loss does not hinder the foreground from changing but rather encourages the foreground to change in a consistent way. In this work, we apply a masked PMSE loss for a single foreground object because of the nature of our data, but one can trivially extend this to multiple foreground objects.
\section{Evaluation}
\label{sec:experiments}

\begin{figure}[t]
\includegraphics[width=\linewidth]{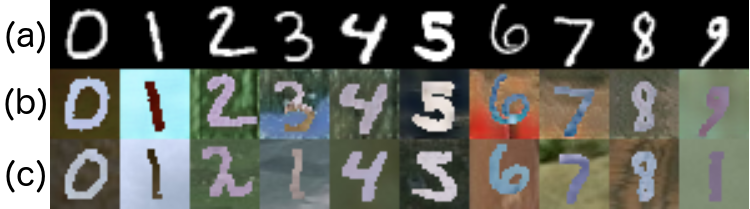}
\caption{Visualization of our model's ability to generate samples when trained
to adapt MNIST to MNIST-M. (a) Source images ${\bs x}^s$ from MNIST;
(b) The samples adapted with our model $G({\bs x}^s, {\bs z})$ with
random noise ${\bs z}$;
(c) The nearest neighbors in the MNIST-M training set of the generated samples in the middle
row. Differences between the middle and bottom rows suggest that the model is not memorizing the target dataset.}
\label{fig:mnist-m_results}
\vspace{-12 pt}
\end{figure}

We evaluate our method on object classification datasets used in previous work\footnote{The most commonly used dataset for visual domain adaptation in the context of object classification is Office \cite{saenko2010adapting}. However, we do not use it in this work as there are significant high--level variations due to label pollution. For more information, see the relevant explanation in \cite{bousmalis2016domain}.},
including MNIST, {MNIST-M} \cite{ganin2016domain}, and USPS \cite{denker1988neural}
as well as a variation of the
LineMod dataset~\cite{hinterstoisser2012accv,wohlhart2015learning}, a standard for object instance
recognition and 3D pose 
estimation, for which we have
synthetic and real data.
Our evaluation is composed of qualitative and quantitative components, using a number of unsupervised domain adaptation scenarios. The qualitative evaluation involves the examination of the ability of our method to learn the underlying pixel adaptation process from the source to the target domain by visually inspecting the generated images. The quantitative evaluation involves a comparison of the performance of our model to previous work and to ``Source Only'' and ``Target Only'' baselines that do not use any domain adaptation. In the first case, we train models only on the unaltered 
source training data and evaluate on the target test data. In the ``Target Only'' case we train task models on the target domain training set only and evaluate
on the target domain test set.
The unsupervised domain adaptation scenarios we consider are listed below: 
\textbf{MNIST to USPS:} Images of the 10 digits (0-9) from the
MNIST~\cite{lecun1998gradient} dataset are used as the source domain
and images of the same 10 digits from the USPS~\cite{denker1988neural} dataset represent
the target domain. To ensure a fair comparison between the ``Source--Only''
and domain adaptation experiments, we train our models on a subset of 50,000
images from the original 60,000 MNIST training images. The remaining
10,000 images are used as validation set
for the ``Source--Only'' experiment. The standard splits for USPS are used, comprising
of 6,562 training, 729 validation, and 2,007 test images.

\textbf{MNIST to MNIST-M:} MNIST~\cite{lecun1998gradient} digits
represent the source domain and MNIST-M~\cite{ganin2016domain} digits
represent the target domain. MNIST-M is a variation on MNIST
proposed for unsupervised domain adaptation. Its images were created by
using each MNIST digit as a binary mask and inverting with it the colors of
a background image. The background images are random crops uniformly sampled
from the Berkeley Segmentation Data
Set~(BSDS500)~\cite{arbelaez2011contour}. All our experiments
follow the experimental protocol by~\cite{ganin2016domain}.  We use the labels for 1,000 out of the
59,001 MNIST-M training examples
to find optimal hyperparameters.

%\vspace{-0.1cm}
\textbf{Synthetic Cropped LineMod to Cropped LineMod:} The LineMod dataset~\cite{hinterstoisser2012accv} is a dataset of small objects
in cluttered indoor settings imaged in a variety of poses. 
We use a cropped version of the dataset~
\cite{wohlhart2015learning}, where each image is
cropped with one of 11 objects in the center. The 11 objects used are `ape', `benchviseblue', `can', `cat', `driller', `duck', `holepuncher', `iron', `lamp', `phone',  and `cam'.
A second component of the dataset consists of CAD models of these same
11 objects in a large variety of poses rendered on a black background, which we refer to
as \emph{Synthetic Cropped LineMod}. We treat Synthetic Cropped LineMod as
the source dataset and the real Cropped LineMod as the target dataset.
We train our model on 109,208 rendered source images and
9,673 real-world target images for domain adaptation, 1,000 for validation,
and a target domain test set of 2,655 for testing. Using this
scenario, our task involves both classification and pose estimation.
Consequently, our task--specific network $T({\bs x};{\bs \theta}_T) \rightarrow \{{\hat{\bs y}}, \hat{\bs q}\}$ outputs both a class $\hat{\bs y}$ and a 3D pose estimate in the form of a positive unit quaternion vector $\hat{\bs q}$. The task  loss becomes:
\vspace{-3 pt}
\begin{align}
{\cal L}_{t}&(G, T) \; = \; \nonumber \\
&\mathbb{E}_{{\bs x}^s,{\bs y}^s, {\bs z}}\Big[-\bs y^{s^\top} \log{\hat{\bs y}}^s -\bs y^{s^\top} \log{\hat{\bs y}}^f+ \nonumber \\
&\quad\xi \log\left(1-\left|{\bs {q}^s}^\top  {\hat{\bs q}}^s\right|\right)+\xi\log\left(1-\left|{\bs {q}^s}^\top  {\hat{\bs q}}^f\right|\right)\Big]
\end{align}
where the
first and second terms are the classification loss, similar to \autoref{eq:task_loss},
and the third and fourth terms are the log of a 3D rotation metric for
quaternions~\cite{huynh2009metrics}. $\xi$ is the weight for the pose
loss, ${\bs q}^s$ represents the
ground truth 3D pose of a sample, $\{{\hat{\bs y}}^s, \hat{\bs q}^s\}=T({\bs x}^s;{\bs \theta}_T)$, ${\{{\hat{\bs y}}^f, \hat{\bs q}^f\}=T(G({\bs x}^s,{\bs z};{\bs \theta}_G);{\bs \theta}_T)}$.  \autoref{tab:pose_results} reports the mean angle the object would
need to be rotated (on a fixed 3D axis) to move from predicted to
ground truth pose~\cite{hinterstoisser2012accv}.

\begin{figure*}[ht]
\centering
\includegraphics[width=\linewidth]{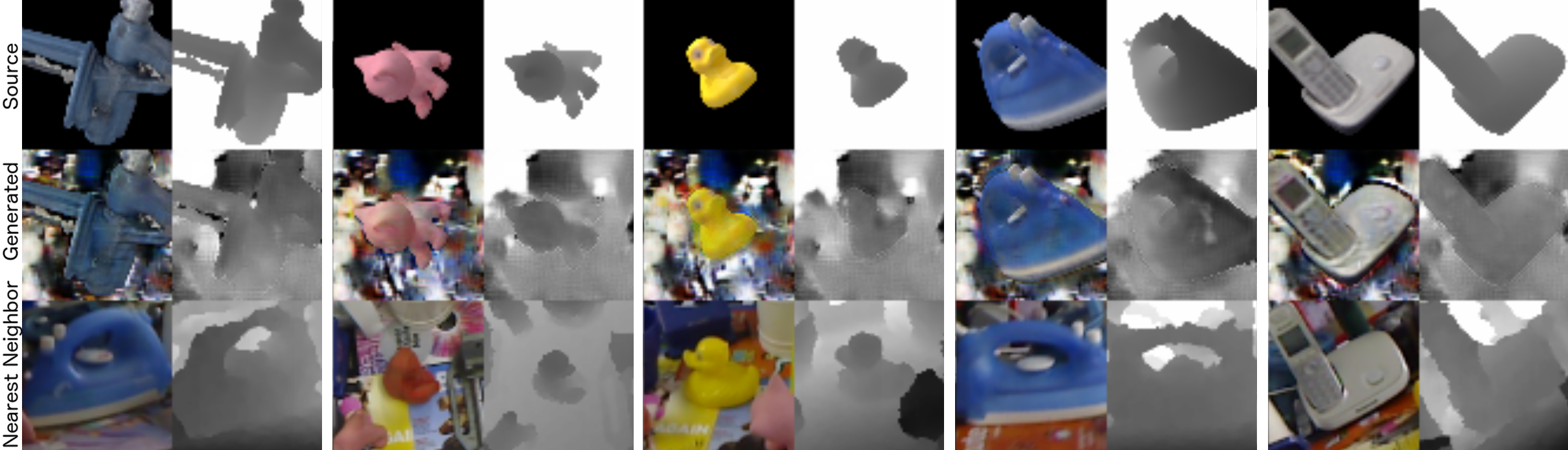}    
\caption{Visualization of our model's ability to generate samples when trained
to adapt Synth Cropped Linemod to Cropped Linemod. \textsl{Top Row:} Source RGB and 
Depth image pairs from Synth Cropped LineMod ${\bs x}^s$; \textsl{Middle Row:} The
samples adapted with our model $G({\bs x}^s, {\bs z})$ with random noise ${\bs z}$;
\textsl{Bottom Row:} The nearest neighbors between the generated samples in the middle
row and images from the target training set. Differences between the generated and
target images suggest that the model is not memorizing the target dataset.}
\label{fig:linemod_results}
\vspace{-12 pt}
\end{figure*}

\vspace{-5 pt}

\subsection{Implementation Details}
All the models are implemented using TensorFlow\footnote{Our code is available here: \mbox{\url{https://goo.gl/fAwCPw}}}~\cite{abadi2016tensorflow} and are trained with the Adam optimizer~\cite{kingma2014adam}. We optimize the objective in \autoref{eq:objective1} for ``MNIST to USPS'' and ``MNIST to MNIST-M'' scenarios and the one in \autoref{eq:objective2} for the ``Synthetic Cropped Linemod to Cropped Linemod'' scenario. We use batches of 32 samples from each domain and the input images are zero-centered and rescaled to $[-1, 1]$.
In our implementation, we let $G$ take the form of a convolutional residual neural network that maintains the resolution of the original image as shown in
Figure~\ref{fig:arch}.  $\bs z$ is a vector of $N^z$ elements, each sampled from a uniform distribution $z_i\sim {\cal U}(-1,1)$. It is fed to a fully connected layer which transforms it to a channel of the same resolution as that of the image channels, and is subsequently concatenated to the input as an extra channel. 
%It is not necessary to use $\bs z$, in which case the generator $G(\bs x^s)$ is deterministic and the only source of stochasticity in the generation process stems from the source domain data distribution.
In all our experiments we use a $\bs z$ with $N^z=10$. The discriminator $D$ is a convolutional neural network where the number of layers depends on the image resolution:  the first layer is a stride 1x1 convolution (motivated by \cite{odena2016deconvolution}), which is followed by repeatedly stacking stride 2x2 convolutions until we reduce the resolution to less or equal to 4x4. The number of filters is $64$ in all layers of $G$, and is 64 in the first layer of $D$ and repeatedly doubled in subsequent layers. The output of this pyramid is fed to a fully--connected layer with a single activation for the domain classification loss. \footnote{Our architecture details can be found in the supplementary \mbox{material}.} 
For all our experiments, the CNN topologies used for the task classifier $T$ are identical to the ones used in ~\cite{ganin2016domain,bousmalis2016domain} to be comparable to previous work in unsupervised domain adaptation.

\subsection{Quantitative Results}

\begin{table}[t]
\centering
\caption{Mean classification accuracy (\%) for digit datasets. The ``Source-only'' and ``Target-only'' rows are the results on the target domain when  using no domain adaptation and training only on the source or the target domain respectively. We note that our Source and
Target only baselines resulted in different numbers than previously published works which
we also indicate in parenthesis.}
\begin{tabular}{| l | c | c |}
\hline
\multirow{2}{*}{\bf Model} & \bf MNIST to&\bf MNIST to\\
&\bf USPS &\bf MNIST-M\\ \hline \hline
Source Only  & 78.9 &  63.6 (56.6) \\ \hline \hline
CORAL \cite{sun2015return} & 81.7 & 57.7 \\ \hline
MMD  \cite{tzeng2015ddc,long2015learning}  & 81.1 & 76.9 \\ \hline 
DANN \cite{ganin2016domain}  & 85.1 & 77.4  \\ \hline
DSN \cite{bousmalis2016domain} & 91.3 & 83.2 \\ \hline
CoGAN \cite{liu2016coupled} & 91.2 & 62.0 \\ \hline
%DRCN \cite{ghifary2016deep} & 91.8 & NA \\ \hline
Our PixelDA & \bf{95.9} & \bf{98.2} \\ \hline
\hline
Target-only  & 96.5 & 96.4 (95.9) \\ \hline
\end{tabular}
\label{tab:results}
\vspace{-17 pt}
\end{table}

We have not found a universally applicable way to optimize hyperparameters for unsupervised domain adaptation.  
Consequently, we follow the experimental protocol of \cite{bousmalis2016domain} and use a small set ($\sim$1,000) of labeled target domain data as a validation set for the hyperparameters of all the methods we compare. We perform all experiments using the same protocol to ensure fair and meaningful comparison. The performance on this validation set can serve as an \textsl{upper bound} of a satisfactory validation metric for unsupervised domain adaptation. As we discuss in section \ref{sec:semi-supervised}, we also evaluate our model in a semi-supervised setting with 1,000 labeled examples in the target domain, to confirm that PixelDA is still able to improve upon the naive approach of training on this small set of target labeled examples. 

We evaluate our model using the aforementioned combinations of source
and target datasets, and compare the performance of our model's task architecture $T$ to that of other state-of-the-art unsupervised domain adaptation techniques based on the same task architecture $T$. As mentioned above, in order to evaluate the efficacy of our model, we first 
compare with the accuracy of models trained in a ``Source Only'' setting for each domain adaptation scenario. This setting represents a lower bound on 
performance.
Next we compare models in a ``Target Only'' setting for each scenario.
This setting represents a weak upper bound
on performance---as it is conceivable that a good unsupervised domain adaptation model
might improve on these results, as we do in this work for ``MNIST to MNIST-M''.

Quantitative results of these comparisons are presented in Tables~\ref{tab:results} and \ref{tab:pose_results}. Our method is able to not just achieve better results than previous work on the ``MNIST to \mbox{MNIST-M}'' scenario, it is also able to outperform the ``Target Only'' performance we are able to get with the same task classifier. Furthermore, we are also able to achieve state-of-the art results for the ``MNIST to USPS'' scenario.
Finally, PixelDA is able to reduce the mean angle error for the ``Synth Cropped Linemod to Cropped Linemod'' scenario to more than half compared to the previous state-of-the-art.

\vspace{-5 pt}
\subsection{Qualitative Results}
\vspace{-3 pt}
The qualitative results of our model are illustrated in figures~\ref{fig:teaser}, \ref{fig:mnist-m_results}, and \ref{fig:linemod_results}. In figures \ref{fig:mnist-m_results} and \ref{fig:linemod_results} one can see the visualization of the generation process, as well as the nearest neighbors of our generated samples in the target domain. In both scenarios, it is clear that our method is able to learn the underlying transformation process that is required to adapt the original source images to images that look like they could belong in the target domain. As a reminder, the MNIST-M digits have been generated by using MNIST digits as a binary mask to invert the colors of a background image. It is clear from figure \ref{fig:mnist-m_results} that in the ``MNIST to MNIST-M'' case, our model is able to not only generate backgrounds from different noise vectors $\bs z$, but it is also able to learn this inversion process. This is clearly evident from e.g. digits $\bs 3$ and $\bs 6$ in the figure. In the ``Synthetic Cropped Linemod to Cropped Linemod'' case, our model is able to sample, in the RGB channels, realistic backgrounds and adjust the photometric properties of the foreground object. In the depth channel it is able to learn a plausible noise model. 

\begin{table}[h]
\centering
\caption{Mean classification accuracy and pose error for the ``Synth Cropped Linemod to Cropped Linemod'' scenario. }
\label{tab:pose_results}
\begin{tabular}{|l|c|c|}
\hline
\multirow{2}{*}{\bf Model} & \bf Classification &\bf Mean Angle\\
&\bf Accuracy &\bf Error\\
\hline \hline
Source-only  & 47.33\%  & $89.2^{\circ}$ \\ \hline  \hline
MMD \cite{tzeng2015ddc,long2015learning} & 72.35\%  & $70.62^{\circ}$ \\ \hline
DANN \cite{ganin2016domain} & 99.90\% &$56.58^{\circ}$ \\ \hline
DSN \cite{bousmalis2016domain} & \textbf{100.00}\% & $53.27^{\circ}$ \\ \hline
Our PixelDA & 99.98\% & $\textbf{23.5}^{\circ}$ \\ \hline \hline
Target-only & 100.00\% & $6.47^{\circ}$ \\ 
\hline
\end{tabular}
\vspace{-5 pt}
\end{table}

% MNIST-M to MNIST
% TR-DA \cite{domaintransduction} & 86.7 & NA & 83.9 \\ \hline
% 
% Ours:  94.03

\vspace{-5 pt}
\subsection{Model Analysis}
\vspace{-3 pt}
\label{sec:model_analysis}
We present a number of additional experiments that demonstrate how the model
works and to explore potential limitations of the model.

\begin{table}[t]
\centering
\caption{Mean classification accuracy and pose error when varying the background of images from the source domain. For these experiments we used only the {RGB} portions of the images, as there is no trivial or typical way with which we could have added backgrounds to depth images. For comparison, we display results with black backgrounds and Imagenet backgrounds (INet), with the ``Source Only'' setting and with our model for the RGB-only case.}
\vspace{6 pt}
\label{tab:background_exp}
\begin{tabular}{|l|c|c|}
\hline
\multirow{2}{*}{\bf Model--RGB-only} & \bf Classification &\bf Mean Angle\\
&\bf Accuracy &\bf Error\\
\hline \hline
Source-Only--Black & $47.33\%$ & $89.2^{\circ}$ \\ \hline
PixelDA--Black & $94.16\%$ & $55.74^{\circ}$  \\ \hline %404T9
Source-Only--INet & $91.15\%$ & $50.18^{\circ}$ \\ \hline
PixelDA--INet & $96.95\%$ & $36.79^{\circ}$ \\ \hline %1838T2
\end{tabular}
\vspace{-9 pt}
\end{table}

\vspace{-12 pt}
\paragraph{Sensitivity to Used Backgrounds}

In both the ``MNIST to MNIST-M'' and ``Synthetic-Cropped LineMod to Cropped LineMod'' scenarios,
the source domains are images of digits or objects on black backgrounds.
Our quantitative evaluation (Tables ~\ref{tab:results} and ~\ref{tab:pose_results}) illustrates
the ability of our model to adapt the source images to the target domain style but raises two questions: Is it important that the backgrounds
of the source images are black and how successful are data-augmentation
strategies that use a randomly chosen background image instead?
To that effect we ran additional experiments where we substituted various backgrounds in place of the default
black background for the Synthetic Cropped Linemod dataset. The backgrounds are randomly selected crops of  images from the ImageNet dataset. In these experiments we used only the RGB portion of the images ---for both source and target domains---
since we don't have equivalent ``backgrounds'' for the depth channel.
%Such approaches are relatively common in domain adaptation approaches \cite{chen2016synthesizing} \cite{tatarchenko2016multi}. 
As demonstrated in Table~\ref{tab:background_exp}, PixelDA is able to improve upon training `Source-only' models on source images of objects on either black or random Imagenet backgrounds. 
%This is consistent
% with the observations of prior work in which the use of random backgrounds
% only worked when coupled with a high degree of manual tuning to ensure
% that the lighting of the backgrounds and objects were consistent and the
% object boundary discontinuities did not appear overly synthetic.

\vspace{-12 pt}
\paragraph{Generalization of the Model}

Two additional aspects of the model are relevant to understanding its 
performance. Firstly, is the model actually learning a successful pixel-level data adaptation process, or is it simply memorizing the target images and replacing
the source images with images from the target training set? Secondly,
is the model able to generalize about the two domains in a fashion not
limited to the classes of objects seen during training?

To answer the first question, we first run our generator $G$ on images from the
source images to create an \emph{adapted} dataset. Next, for each
transferred image, we perform a pixel-space $L2$ nearest neighbor lookup in the target
training images to determine whether the model is simply memorizing images
from the target dataset or not. Illustrations are shown in 
figures~\ref{fig:mnist-m_results} and \ref{fig:linemod_results}, where the top rows are samples from ${\bs x}^s$, the middle rows are generated samples $G({\bs x}^s, {\bs z})$, and the bottom rows are the nearest neighbors of the generated samples in the target training set. It is clear from the figures that the model is not memorizing images from the target training set.

Next, we evaluate our model's ability to generalize to classes unseen
during training. To do so, we retrain our best model using a subset of images
from the source and target domains which includes only half of the object 
classes for the ``Synthetic Cropped Linemod'' to ``Cropped Linemod'' scenario. Specifically, the objects `ape', `benchviseblue', `can', `cat', `driller', and `duck' are observed during the training procedure, and the other objects are only used during testing. Once  $G$ is trained, we fix its weights and pass the full training set of the source domain to generate images used for training the task-classifier $T$. We then evaluate the performance of $T$ on the entire set of unobserved objects (6,060 samples), and the test set of the target domain for all objects for direct comparison with Table ~\ref{tab:pose_results}.

\begin{table}[t]
\centering
\caption{Performance of our model trained on only 6 out of 11 Linemod objects. The first row, `Unseen Classes,' displays the performance on all the samples of the remaining 5 Linemod objects not seen during training. The second row, `Full test set,' displays the performance on the target domain test set for all 11 objects. }
\label{tab:label_space_exp}
\begin{tabular}{|l|c|c|}
\hline
\multirow{2}{*}{\bf Test Set} & \bf Classification &\bf Mean Angle\\
&\bf Accuracy &\bf Error\\
\hline \hline
Unseen Classes & $98.98\%$ & $31.69^{\circ}$  \\ \hline
Full test set & $99.28\%$ & $32.37^{\circ}$ \\\hline
\end{tabular}
\vspace{-6 pt}
\end{table}

\vspace{-15 pt}
\paragraph{Stability Study}

We also evaluate the importance of the different components of our model.
We demonstrate that while the task and content losses do not improve the overall
performance of the model, they dramatically stabilize training. Training instability is a common
characteristic of adversarial training, necessitating various strategies to deal with
model divergence and mode collapse \cite{salimans2016improved}. We measure the standard deviation of the performance of our models by running each model $10$ times with different random parameter initialization but with the same hyperparameters. Table~\ref{tab:stability} illustrates that the use of the task and content--similarity losses reduces the level of variability across runs.

\begin{table}[t]
\centering
\caption{The effect of using the task and content losses  $L_{t}$, $L_{c}$ on the standard deviation (std) of the performance of our model on the ``Synth Cropped Linemod to Linemod'' scenario. $L_{t}^{source}$ means we use source data to train $T$; $L_t^{adapted}$ means we use generated data to train $T$; $L_c$ means we use our content--similarity loss. A lower std on the performance metrics means that the results are more easily reproducible.}
\label{tab:stability}
\begin{tabular}{|c|c|c|c|c|}
\hline
\multirow{2}{*}{$L_{t}^{source}$} & \multirow{2}{*}{$L_{t}^{adapted}$} & \multirow{2}{*}{$L_{c}$} & \bf{Classification} & \bf{Mean Angle}\\
&&&\bf{Accuracy std} & \bf{Error std}\\\hline
-&-&-& 23.26 & 16.33\\\hline
-&\checkmark&-&  22.32  & 17.48 \\ \hline
\checkmark&\checkmark&-&  2.04 & \bf 3.24\\\hline
\checkmark&\checkmark&\checkmark&  \bf 1.60 & 6.97\\
\hline
\end{tabular}
\vspace{-9 pt}
\end{table}

\subsection{Semi-supervised Experiments}
\label{sec:semi-supervised}
Finally, we evaluate the usefulness of our model in a semi--supervised setting, in which we assume we have a small number of labeled target training examples. The semi-supervised version of our model simply uses these additional training samples as extra input to classifier $T$ during training. We sample 1,000 examples from the Cropped Linemod not used in any previous experiment and use them as additional training data. We evaluate the semi-supervised version of our model on the test set of the Cropped Linemod target domain against the 2 following baselines: (a) training a classifier only on these 1,000 target samples without any domain adaptation, a setting we refer to as `1,000-only'; and (b) training a classifier on these 1,000 target samples and the entire Synthetic Cropped Linemod training set with no domain adaptation, a setting we refer to as `Synth+1000'. As one can see from Table ~\ref{tab:semi-supervised} our model is able to greatly improve upon the naive setting of incorporating a few target domain samples during training. We also note that PixelDA leverages these samples to achieve an even better performance than in the fully unsupervised setting (Table ~\ref{tab:pose_results}).

\vspace{-10 pt}
\begin{table}[t]
\centering
\caption{Semi-supervised experiments for the ``Synthetic Cropped Linemod to Cropped Linemod'' scenario. When a small set of 1,000 target data is available to our model, it is able to improve upon baselines trained on either just these 1,000 samples or the synthetic training set augmented with these labeled target samples.}
\label{tab:semi-supervised}
\begin{tabular}{|l|c|c|}
\hline
\multirow{2}{*}{\bf Method} & \bf Classification &\bf Mean Angle\\
&\bf Accuracy &\bf Error\\
\hline \hline
1000-only & $99.51\%$ & $25.26^{\circ}$ \\ \hline
Synth+1000 & $99.89\%$ & $23.50^{\circ}$  \\ \hline
Our PixelDA & $\bf 99.93\%$ & $\bf 13.31^{\circ}$ \\\hline
\end{tabular}
\vspace{-14 pt}
\end{table}

\section{Conclusion}
\label{sec:conclusion}
\vspace{-3 pt}
We present a state-of-the-art method for performing unsupervised domain adaptation. Our PixelDA models outperform previous work on a set of unsupervised domain adaptation scenarios, and in the case of the challenging ``Synthetic Cropped Linemod to Cropped Linemod'' scenario, our model more than halves the error for pose estimation compared to the previous best result. They are able to do so by using a GAN--based technique, stabilized by both a task-specific loss and a novel content--similarity loss. Furthermore, our model decouples the process of domain adaptation from the task-specific architecture, and provides the added benefit of being easy to understand via the visualization of the adapted image outputs of the model.
{
\paragraph*{Acknowledgements}
The authors would like to thank Luke Metz, Kevin Murphy, Augustus Odena, Ben Poole, Alex \mbox{Toshev}, and Vincent Vanhoucke for suggestions on early drafts of the paper.
}
{\small
\bibliographystyle{ieee}
\bibliography{synth_to_real}
}
\onecolumn
\newpage
\renewcommand\thesubsection{\Alph{subsection}}
\renewcommand\thesubsubsection{\Alph{subsection}.\arabic{subsubsection}}
\setcounter{figure}{0}
\setcounter{section}{0}
\setcounter{subsection}{0}
\pagenumbering{arabic}

\section*{Supplementary Material}

\subsection{Additional Generated Images}

\begin{figure*}[h]
\centering
\includegraphics[width=\textwidth]{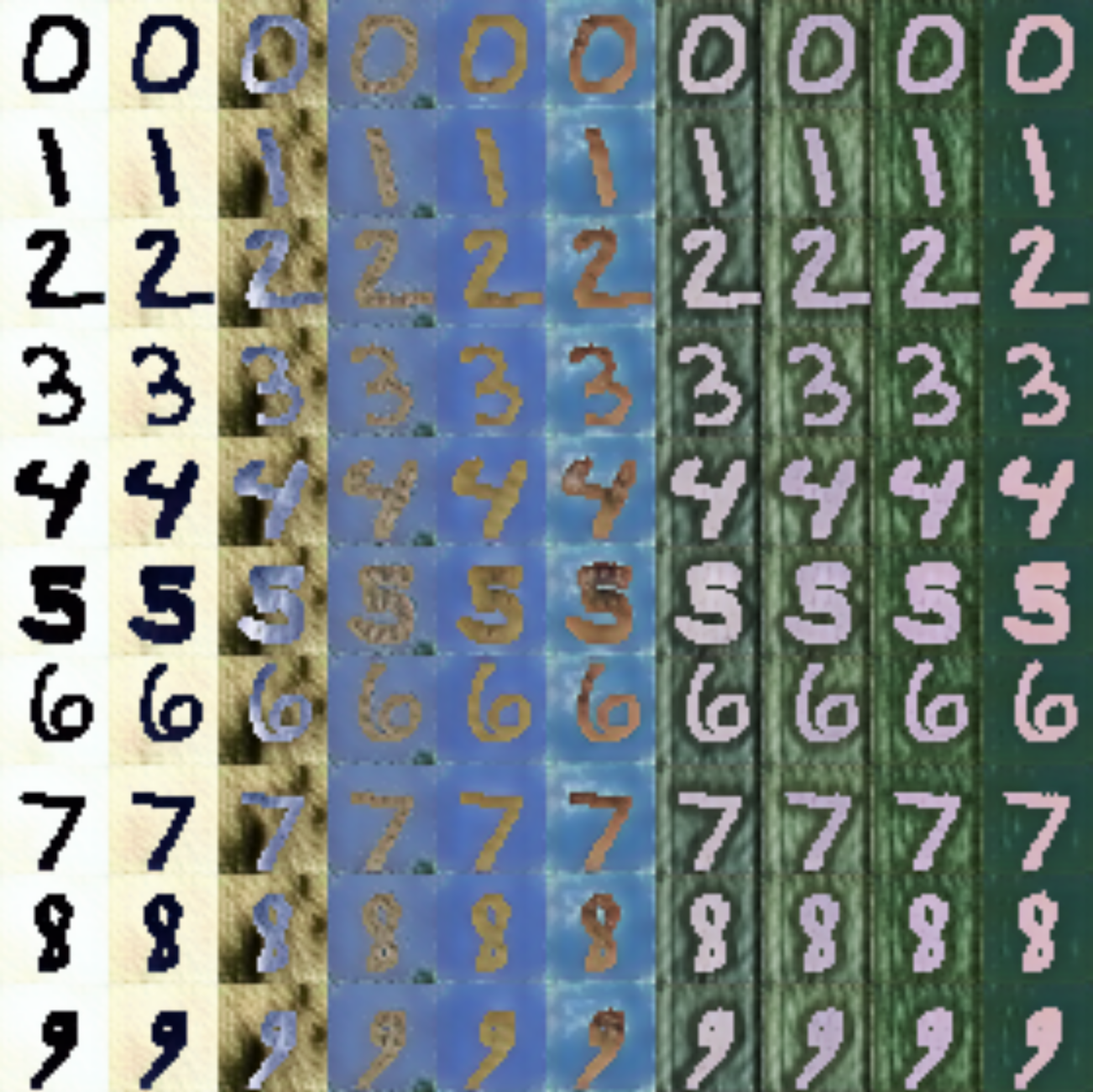}
\caption{Linear interpolation between two random noise vectors demonstrates that the model is able to separate out style from content in the MNIST-M dataset.  Each row is generated from the same MNIST digit, and each column is generated with the same noise vector.}
\end{figure*}

\begin{figure*}[h]
\centering
\begin{subfigure}{0.4\textwidth}
\includegraphics[width=\textwidth,right]{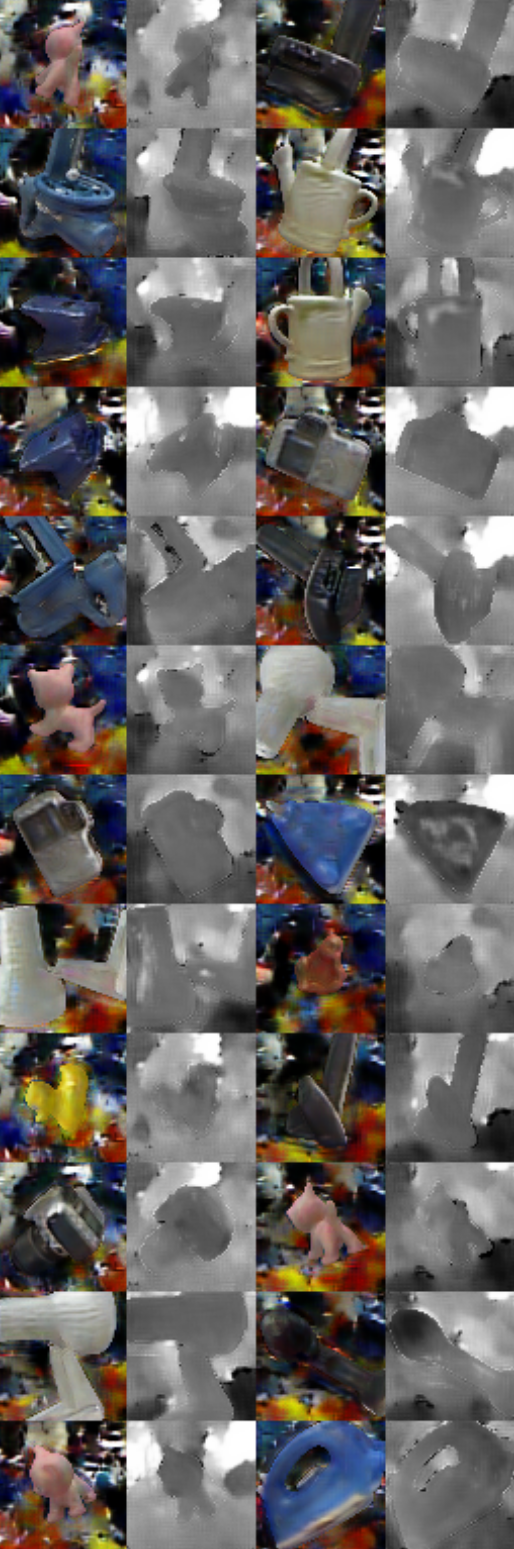}
\end{subfigure}
\begin{subfigure}{0.4\textwidth}
\includegraphics[width=\textwidth,left]{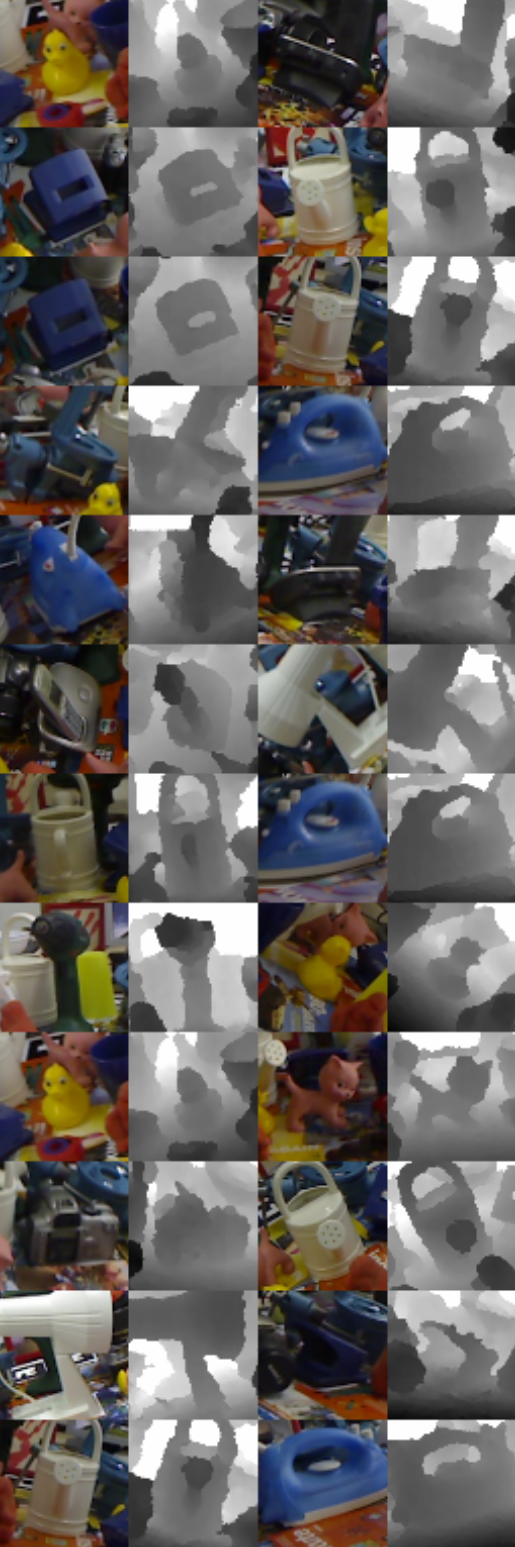}
\end{subfigure}
\caption{Additional generation examples for the LineMod dataset. The left 4 columns are generated images and depth channels, while the corresponding right 4 columns are L2 nearest neighbors.}
\end{figure*}

\newpage
%\FloatBarrier
\subsection{Model Architectures and Parameters}
\begin{figure}[b]
\includegraphics[width=\textwidth]{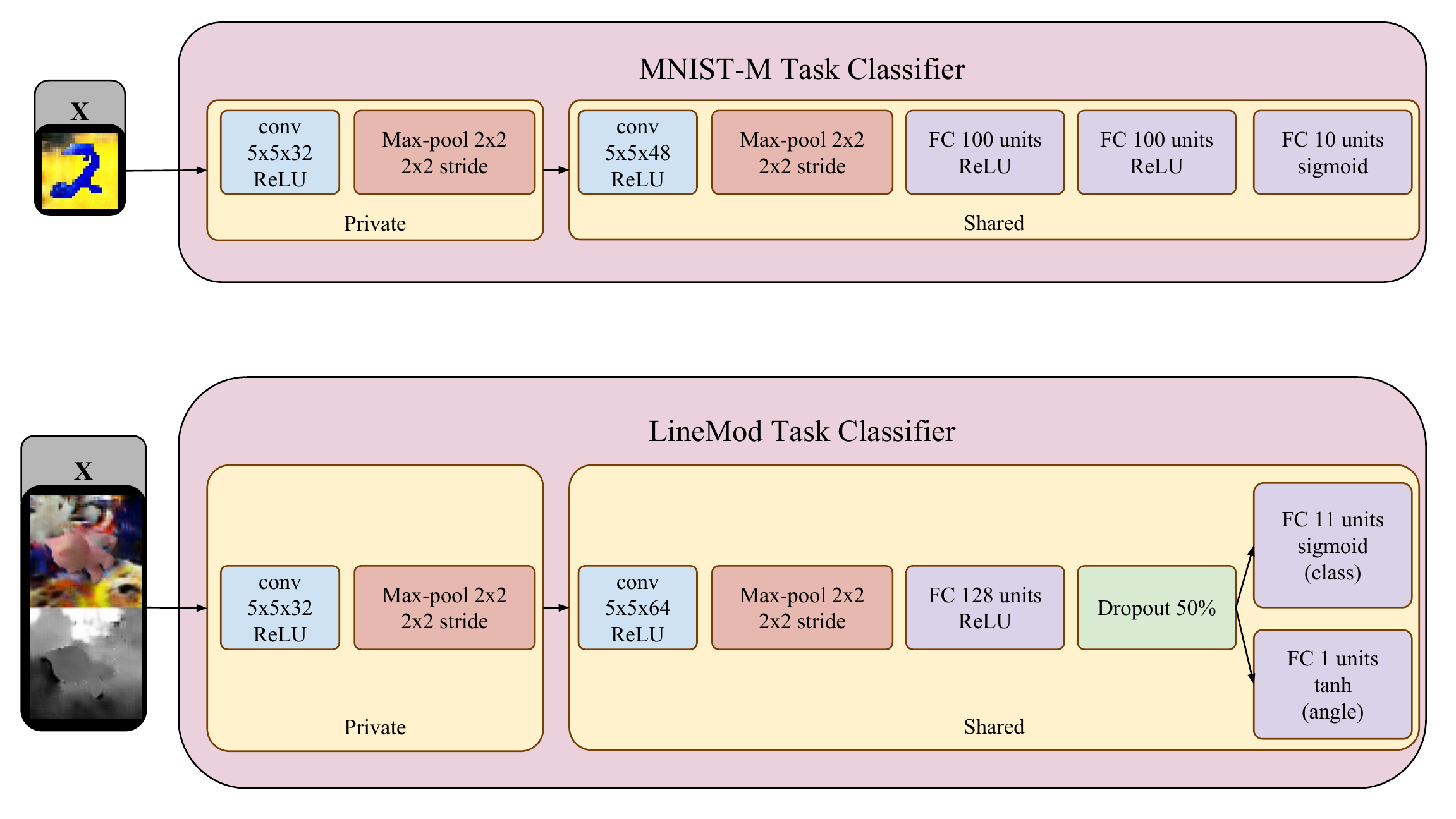}
\caption{Task classifier (T) architectures for each dataset. We use the same task classifiers as ~\cite{bousmalis2016domain,ganin2016domain} to enable fair comparisons.  The MNIST-M classifier is used for USPS, MNIST, and MNIST-M classification.
During training, the task classifier is applied to both the synthetic and generated images when $L_t^{source}$ is enabled (see Paper \autoref{tab:stability}).
The 'Private' parameters are only trained on one of these sets of images, while the 'Shared' parameters are shared between the two.  At test time on the target domain images, the classifier is composed of the Shared parameters and the Private parameters which are part of the generated images classifier.  These first private layers allow the classifier to share high level layers even when the synthetic and generated images have different channels (such as 1 channel MNIST and RGB MNIST-M).}
\end{figure}

We present the exact model architectures used for each experiment along with hyperparameters needed to reproduce results.  The general form for the generator, G, and discriminator, D, are depicted in \autoref{fig:arch} of the paper.  For G, we vary the number of filters and the number of residual blocks. For D, we vary the amount of regularization and the number of layers.

Optimization consists of alternating optimization of the discriminator and task classifier parameters, referred to as the D step, with optimization of the generator parameters, referred to as the G step.

Unless otherwise specified, the following hyperparameters apply to all experiments:

\begin{itemize}
\itemsep0em 
\item Batch size 32
\item Learning rate decayed by 0.95 every 20,000 steps
\item All convolutions have a 3x3 filter kernel
\item Inject noise drawn from a zero centered Gaussian with stddev 0.2 after every layer of discriminator
\item Dropout every layer in discriminator with keep probability of 90\%
\item Input noise vector is 10 dimensional sampled from a uniform distribution ${\cal U}(-1,1)$
\item We follow conventions from the DCGAN paper \cite{RadfordMC15} for several aspects
\begin{itemize}
\itemsep0em 
\item An L2 weight decay of $1e^{-5}$ is applied to all parameters
\item Leaky ReLUs have a leakiness parameter of 0.2
\item Parameters initialized from zero centered Gaussian with stddev 0.02
\item We use the ADAM optimizer with $\beta_1 = 0.5$
\end{itemize}
\end{itemize}

\subsubsection{USPS Experiments}
The Generator and Discriminator are identical to the MNIST-M experiments.  
\vspace{-0.5cm}
\paragraph{Loss weights:}
\begin{itemize}
\itemsep0em 
\item Base learning rate is $2e^{-4}$
\item The discriminator loss weight is 1.0
\item The generator loss weight is 1.0 
\item The task classifier loss weight in G step is 1.0
\item There is no similarity loss between the synthetic and generated images
\end{itemize}

\subsubsection{MNIST-M Experiments (Paper \autoref{tab:results})}
\paragraph{Generator:} The generator has 6 residual blocks with 64 filters each
\vspace{-0.5cm}
\paragraph{Discriminator:} The discriminator has 4 convolutions with 64, 128, 256, and 512 filters respectively.  It has the same overall structure as paper \autoref{fig:arch}
\vspace{-0.5cm}
\paragraph{Loss weights:}
\begin{itemize}
\itemsep0em 
\item Base learning rate is $1e^{-3}$
\item The discriminator loss weight is 0.13
\item The generator loss weight is 0.011 
\item The task classifier loss weight in G step is 0.01
\item There is no similarity loss between the synthetic and generated images
\end{itemize}

\subsubsection{LineMod Experiments}
All experiments are run on a cluster of 10 TensorFlow workers.  We benchmarked the inference time for  the domain transfer on a single K80 GPU as 30 ms for a single example (averaged over 1000 runs) for the LineMod dataset.

\vspace{-0.5cm}
\paragraph{Generator:} The generator has 4 residual blocks with 64 filters each
\vspace{-0.5cm}
\paragraph{Discriminator:} The discriminator matches the depiction in paper \autoref{fig:arch}.  The dropout keep probability is set to 35\%.
\vspace{-0.5cm}
\paragraph{Parameters without masked loss (Paper \autoref{tab:pose_results}):}
\begin{itemize}
\itemsep0em 
\item Base learning rate is $2.2e^{-4}$, decayed by 0.75 every 95,000 steps
\item The discriminator loss weight is 0.004
\item The generator loss weight is 0.011 
\item The task classification loss weight is 1.0
\item The task pose loss weight is 0.2
\item The task classifier loss weight in G step is 0
\item The task classifier is not trained on synthetic images
\item There is no similarity loss between the synthetic and generated images
\end{itemize}

\vspace{-0.5cm}
\paragraph{Parameters with masked loss (Paper \autoref{tab:stability}):}
\begin{itemize}
\itemsep0em 
\item Base learning rate is $2.6e^{-4}$, decayed by 0.75 every 95,000 steps.
\item The discriminator loss weight is 0.0088
\item The generator loss weight is 0.011 
\item The task classification loss weight is 1.0
\item The task pose loss weight is 0.29
\item The task classifier loss weight in G step is 0
\item The task classifier is not trained on synthetic images
\item The MPSE loss weight is 22.9
\end{itemize}

%\FloatBarrier
\subsection{InfoGAN Connection}

In the case that $T$ is a classifier, we can show that optimizing the task loss in the way described in the main text amounts to a variational
approach to maximizing mutual information ~\cite{agakov2004algorithm}, akin to the InfoGAN model ~\cite{chen2016infogan}, between the predicted class and both the generated and the equivalent source images. The classification loss could be re-written, using conventions from ~\cite{chen2016infogan} as:
\begin{align}
{\cal L}_{t} \; &= \; -\mathbb{E}_{{\bs x}^s \sim {\cal D}^s} [ \mathbb{E}_{{y}' \sim p({y}|{\bs x}^s)}\log q({ y}'|{\bs x}^s)]  \nonumber\\
&\quad -\mathbb{E}_{{\bs x}^f \sim G({\bs x}^s, {\bs z})} [ \mathbb{E}_{{y}' \sim p({y}|{\bs x}^f)}\log q({ y}'|{\bs x}^f)]\\
&\ge -I(y', {\bs x}^s) - I(y', {\bs x}^f)+ 2H(y),
\end{align}
where $I$ represents mutual information, $H$ represents entropy, $H(y)$ is assumed to be constant as in ~\cite{chen2016infogan}, $y'$ is the random variable representing the class, and $q({y}'|.)$ is an approximation of the posterior distribution $p({y}'|.)$ and is expressed in our model with the classifier $T$. Again, notice that we maximize the mutual information of $y'$ and the equivalent source and generated samples. By doing so, we are effectively regularizing the adaptation process to produce images that look similar for each class to the classifier $T$. This helps maintain the original content of the source image and avoids, for example, transforming all objects belonging to one class to look like objects belonging to another.

\subsection{Deep Reconstruction-Classification Networks}
Ghifary \etal \cite{ghifary2016deep} report a result of 91.80\% accuracy on the MNIST $\rightarrow$ USPS domain pair, versus our result of 95.9\%.  We attempted to reproduce these results using their published code and our own implementation, but we were unable to achieve comparable performance.

\end{document}